 \documentclass[final,5p,times,twocolumn, authoryear]{elsarticle}

\usepackage{textcomp}
\usepackage{array,ragged2e}
\usepackage{apalike}
\usepackage{xurl} 

\usepackage{orcidlink}
\usepackage{colortbl}
\usepackage{makecell}
\usepackage{pgfplots}
\pgfplotsset{compat=1.17}
\usepackage{amsmath}
\usepackage{multicol}
\usepackage{multirow}
\usepackage{tabto}
\usepackage{wrapfig}
\usepackage{amssymb}
\usepackage{amsmath}
\usepackage{float}
\usepackage{graphicx}
\usepackage[misc,geometry]{ifsym}

\usepackage{tikz}
\usetikzlibrary{positioning}

\usepackage{textcomp}
\usepackage{array,ragged2e}
\usepackage{pgfplots}
\usepackage{url}
\usepackage{lineno,hyperref}
\hypersetup{
  linkcolor  = red!60!black,
  citecolor  = blue!90!white,
  urlcolor   = violet!60!black,
  colorlinks = true,
}

\usepackage{subcaption}
\usepackage[onehalfspacing]{setspace}
\usepackage{eurosym}

\makeatother
\journal{Computers and Electronics in Agriculture}

\begin{document}

\begin{frontmatter}

\title{Applying Time Series Deep Learning Models to Forecast \\ the Growth of Perennial Ryegrass in Ireland}

\author[add1]{Oluwadurotimi Onibonoje}
\author[add2,add3]{Vuong M. Ngo\corref{cor1}}
\author[add1,add2]{Andrew McCarren}
\author[add4]{Elodie Ruelle}
\author[add4]{Bernadette O'Brien}
\author[add1,add2]{Mark Roantree}

\cortext[cor1]{Corresponding author}

\address[add1]{VISTAMILK, Dublin City University, Dublin,  Ireland}
\address[add2]{Insight Centre for Data Analytics, Dublin City University, Dublin, Ireland}
\address[add3]{Ho Chi Minh City Open University, Ho Chi Minh City, Vietnam}
\address[add4]{Teagasc, Animal \& Grassland Research and Innovation Centre, Co. Cork, Ireland}

\begin{abstract}
Grasslands, constituting the world's second-largest terrestrial carbon sink, play a crucial role in biodiversity and the regulation of the carbon cycle. Currently, the Irish dairy sector, a significant economic contributor, grapples with challenges related to profitability and sustainability. Presently, grass growth forecasting relies on impractical mechanistic models. In response, we propose deep learning models tailored for univariate datasets, presenting cost-effective alternatives. Notably, a temporal convolutional network designed for forecasting Perennial Ryegrass growth in Cork exhibits high performance, leveraging historical grass height data with RMSE of 2.74 and MAE of 3.46. Validation across a comprehensive dataset spanning 1,757 weeks over 34 years provides insights into optimal model configurations. This study enhances our understanding of model behavior, thereby improving reliability in grass growth forecasting and contributing to the advancement of sustainable dairy farming practices.

\end{abstract}


\end{frontmatter}


\section{Introduction}

Grasslands stand as the world's largest terrestrial ecosystem, serving as a pivotal source of sustenance for livestock. Tackling the escalating demand for meat and dairy products in an environmentally sustainable manner presents a formidable challenge. Encompassing 31.5\% of the Earth's landmass \citep{Latham:2014}, grasslands rank among the most prevalent and widespread vegetation types. Following forests, they emerge as the second-largest terrestrial carbon sink, playing a critical role in regulating the global carbon cycle. Moreover, grasslands contribute significantly to the support of plant and animal biodiversity, further emphasizing their multifaceted ecological importance \citep{ACDBG16}.
    
The dairy sector plays a significant role in Ireland, contributing substantially to the country's income and employment. This is evident from the sector's impressive performance, generating over €5 billion in exports \citep{Irish-FB:2022} and providing support to over 60,000 jobs in the economy \citep{Fitzgerald:2019}. Additionally, grass-based feeding systems present advantages for Irish dairy farmers due to their relatively low cost and their contribution to reducing the carbon footprint of many Irish dairy farms \citep{Brien:16}. Despite being economically and environmentally significant, Irish dairy farmers face challenges in enhancing profitability, preserving cost competitiveness, and adhering to sustainable farming practices.

Forecasting grass growth is crucial for dairy production, considering that grass serves as an environmentally sustainable and cost-effective feed source \citep{Jessica:2017, MMBD21}. Precise measurements of grass growth rates and accurate forecasts are foundational to effective grazing management \citep{HGO17}, guiding resource allocation decisions for grassland farmers and other stakeholders. Additionally, observing the spatio-temporal dynamics of changes in both the quality and quantity of above-ground biomass in grasslands holds significance \citep{Lussem:2019}. This monitoring process proves valuable for adapting management decision-making, including adjustments to stocking density, mowing times, or fertilizer application rates \citep{Viljanen:2018}. Nevertheless, accurately predicting grass growth poses a considerable challenge due to the intricate interactions between factors such as meteorological parameters, soil type and fertility, and nutrient availability \citep{RHD18}.

The prevailing methods for forecasting grass growth predominantly lean on mechanistic models, demanding meticulous tracking of various parameters such as climate, nitrogen availability, and soil conditions. However, the exhaustive nature of this data collection process makes it impractical for numerous farms. This emphasizes the crucial need for rapid and data-efficient alternatives that can maintain robust predictive accuracy. Finding such substitutes is pivotal to ensuring that farms, even those with limited resources, can benefit from effective grass growth forecasting without the burden of extensive data collection efforts. Striking a balance between accuracy and feasibility becomes paramount in enhancing the accessibility and applicability of forecasting methods in diverse agricultural settings.

In summary, our work makes the following contributions:
\begin{itemize}
    \item We proposed deep learning (DL) models for forecasting in a univariate grass growth dataset, requiring minimal overhead in data acquisition. In contrast to other research that often focuses on grass growth prediction using a multivariate approach, which demands a significantly more expensive and resource-intensive data collection effort.

    \item The temporal convolutional network presented in our study showcases its prowess in accurately forecasting the growth of Perennial Ryegrass in Cork, delivering high-performance results. By leveraging past grass height data, this approach provides a robust mechanism for predicting future heights with precision.

    \item In our extensive validation process, we assessed the performance and runtimes of our proposed approach across a comprehensive dataset spanning 1,757 weeks over a 34-year period, from 1982 to 2015. This thorough investigation allowed us to delve into the intricacies of DL models, focusing particularly on determining the optimal configurations such as the number of layers and sequence length. The insights gained from this analysis contribute to a more nuanced understanding of the model's behavior over an extended time frame, further enhancing its applicability and reliability in forecasting grass growth.

\end{itemize}

The remaining sections of the paper are organized as follows. In Section~\ref{sec:relatedwork}, we delve into a review of related work. Following that, Section \ref{sec:dataset} provides comprehensive information about the dataset used for training and evaluating our DL models. Section \ref{sec:models} introduces the DL models designed specifically for forecasting grass height in time series. In Section \ref{sec:exper}, we elaborate on the dataset processing, evaluation metrics, and hyperparameter tuning procedures for our models. Subsequently, Section \ref{sec:results} presents, analyzes, and discusses the experimental results. Finally, we draw conclusions and outline future directions in Section \ref{sec:conclusion}.
    
\section{Related Work} \label{sec:relatedwork}

Several studies have leveraged information technologies in agriculture to enhance crop yield, as exemplified by \cite{Ngo:2021}, \cite{Ngo:2023}, and \cite{Benedict:2023}. In \cite{Ngo:2021}, the authors introduced an Electronic Farming Record (EFR) and utilized agricultural Big Data analytics to determine optimal quantities of various factors, including soil properties (texture and pH), soil nutrients, seed rates, herbicides, insecticides, fungicides, and adjuvants for the 12 most popular crops in Europe. Meanwhile, \cite{Ngo:2023} employed data warehousing and statistical techniques \citep{ngo2019designing} to propose recommended quantities of fertilizer components (such as nitrogen, phosphorus, and potassium) for Barley, Dried Beans, Linseed, Rye, and Wheat, considering a broad spectrum of environmental and crop management conditions. In \cite{Benedict:2023}, the authors studied exotic annual grass (EAG) in western U.S. rangeland during the active growing season. They used a normalized difference vegetation index (NDVI) threshold-based interpolation technique to understand the links between weather conditions, EAG phenology, and the potential impact of weed grass competition on crop yield and quality. However, these papers did not employ time series models, DL models, or provide forecasts for grass yield.

Several papers have utilized machine learning (ML)/DL models for identifying information related to grass various countries, such as \cite{Sapkota:2020}, \cite{Holtgrave:2023}, and \cite{Defalque:2024}. In the study by \cite{Sapkota:2020}, the authors implemented a deep neural network based on an unmanned aerial systems-based remote sensing approach, incorporating color-transformed features and vegetation indices. This amalgamation significantly improved the detection and mapping of Italian ryegrass in wheat. In the study by \citep{Holtgrave:2023}, the authors applied Convolutional Neural Network (CNN) and Long Short-Term Memory (LSTM) models, utilizing optical, synthetic aperture radar, and weather time series data. Their objective was to identify grassland mowing events in Germany. In the study conducted by \citep{Defalque:2024}, ML models, namely the Xtreme Gradient Boosting Regressor and Support Vector Regressor, were formulated and implemented. These models considered a range of factors, encompassing cattle parameters, environmental conditions, and spectral data within the Brazilian study area. The primary aim of these models was to estimate biomass and dry matter in grazing systems. However, these three papers did not use time series ML/DL models nor made forecasts regarding grass height.

Some papers applied ML/DL algorithms to forecast grass growth/height, such as \cite{KRGS19}, \cite{MBLJ20} and \cite{PBADGQ21}. In the study by \cite{KRGS19}, a case-based reasoning system was employed for predicting grass growth, integrating a Bayesian model to account for climatic variability and data uncertainty. The methodology included employing a gold standard dataset as a model to eliminate noise from the working dataset. Cases were generated using features such as the average growth rate since the previous grass cover, the recorded week, month, and season. Additionally, the study incorporated weather data, encompassing maximum temperature, average soil temperature, and average global radiation. In the work by \cite{MBLJ20}, the authors introduced a multi-layered mapping methodological framework aimed at addressing challenges in precision agriculture. This framework serves to enhance transparency and explainability in decision support systems. The paper conducted a preliminary exploratory statistical case study analysis on grass growth data, specifically focusing on Northern Ireland. The analysis aimed to unveil patterns and identify the key factors influencing grass growth in the region. In the study by \cite{PBADGQ21}, the authors utilized ML algorithms to improve the accuracy of predicting dry matter yield for perennial ryegrass. They applied Partial Least Squares Regression, Random Forest, and Support Vector Machines to 468 field plots, incorporating both structural data from a consumer-grade RGB camera and spectral information from a Multispectral camera system. The results revealed that the optimal performance was achieved through the combination of Multispectral information and the Random Forest algorithm. However, these three papers did not employ time series models for forecasting.

There are some papers applied time series techniques in agriculture, such as \cite{Yoo:2020},  \cite{Yuan:2023} and \cite{Quan:2023}. The authors in \cite{Yoo:2020} introduced a LSTM model designed to forecast the sales of agricultural products, aiming to stabilize supply and demand. The model incorporated seasonality attributes such as week, month, and quarter as additional inputs to historical time-series data. The evaluation focused on 3000 items, including crops like Onion, Lettuce, Mallow, and Tomato. These sales records were collected from the point-of-sale system of a local food retail store in Wanju, South Korea, spanning the period from June 2014 to December 2019. In \cite{Yuan:2023}, the authors  introduced contrastive learning as a methodology for extracting unified representations in the context of crop classification, utilizing both optical and synthetic aperture radar satellite image time series. They innovatively devised an enhanced feature-level fusion network with selectively shared weights among branches, aiming to mitigate model complexity. In \cite{Quan:2023}, the authors utilized multi-temporal remote sensing images that integrated both spectral and lidar data. The dataset employed in their research encompasses five distinct periods of maize growth in Harbin, China, during the year 2021. Employing sophisticated deep learning techniques, they conducted a spatial assessment of weed competitiveness across various stages of maize growth. The paper presented the dynamic responses of spectral and lidar information during prolonged weed competition. However, the study's emphasis was on image processing and the time series data covered only a one-year period. However, \cite{Yoo:2020} studied non-grass agricultural products using a dataset spanning only approximately five years. While, \cite{Yuan:2023} worked on image processing and classification. In \cite{Quan:2023}, the study's emphasized image processing, with the time series data covering only a one-year period.



\section{Dataset} \label{sec:dataset}

\begin{figure}[htbp]
    \centering
    \includegraphics[width=8.6cm]{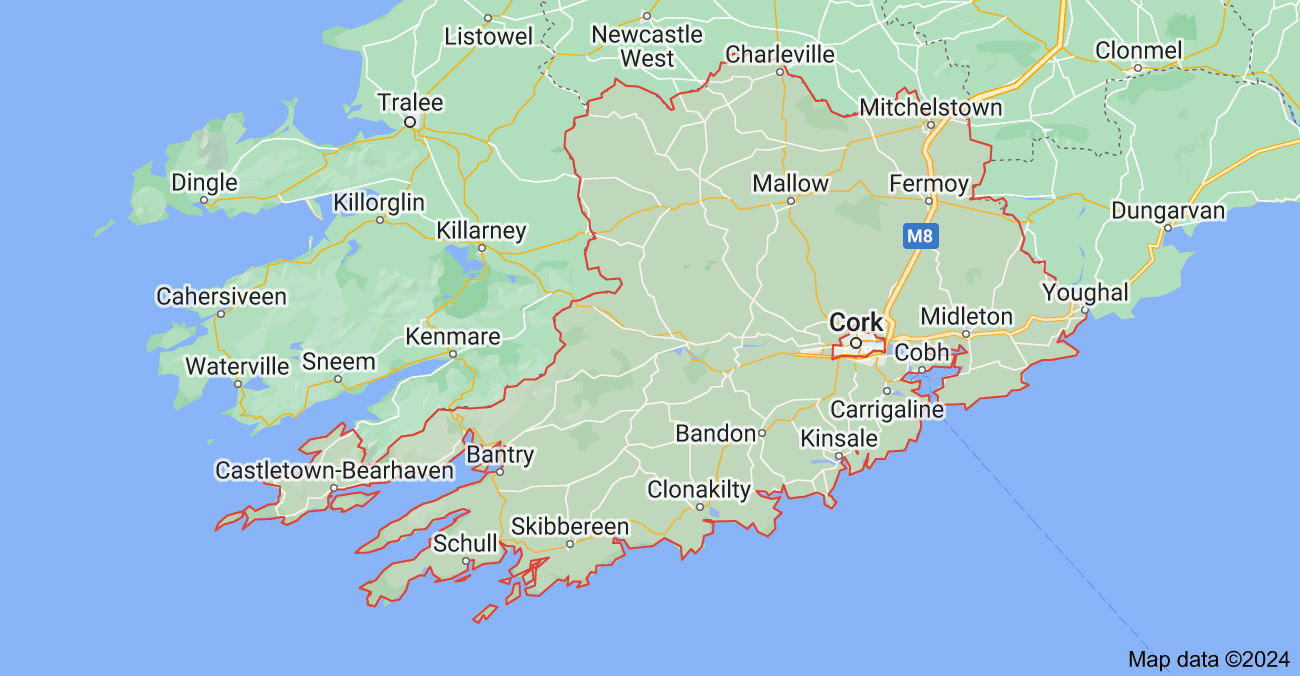}
    \caption{County Cork in the South of Ireland}
    \label{fig:cork_map.png}
\end{figure}

Researchers at the Teagasc, Animal \& Grassland Research and Innovation Centre collected weekly data on grass growth in fields located in County Cork, Ireland, as illustrated in Figure \ref{fig:cork_map.png}. the dataset spans 34 years, ranging from 1982 to 2015. County Cork is marked by moderately warm temperatures, with maximum temperatures surpassing 25 ºC in the summer months, and mild winters, where minimum temperatures typically range from 0 to 5 ºC. The Köppen climate classification subtype assigned to this climate is "Cfb," indicating a Marine West Coast Climate \citep{BZMV18}

\begin{figure}
\scriptsize
\centering
\begin{tikzpicture}
  \node (img)  {\includegraphics[width=8cm]{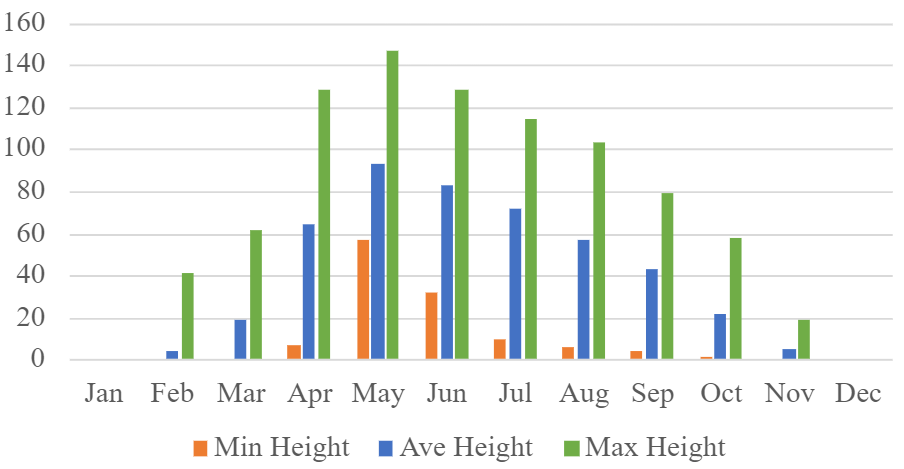}};
  \node[below=of img, node distance=0cm,   yshift=1cm,] {Month in a year};
  \node[left=of img, node distance=0cm, rotate=90, anchor=center, xshift=0.5cm, yshift=-0.8cm,] {Height (cm)};
\end{tikzpicture}
\caption{Monthly Height of Ryegrass}
\label{fig:grass_height_monthly.png}
\end{figure}

Figure \ref{fig:grass_height_weekly.png} displays the weekly recorded growth height recorded weekly over 1,757 weeks. The week with the highest growth is the 223$^{rd}$ week, on 13-May-1986, reaching a height of 146.4 cm. Meanwhile, Figure \ref{fig:grass_height_monthly.png} illustrates the monthly average height of Ryegrass in year. It is evident that grass growth tends to peak from mis-spring to summer (April to August) and begins to decline from fall to winter. The highest and second-highest average heights of the grass are 92.7 cm and 82.9 cm in May and June, respectively.

\begin{figure}
\scriptsize
\begin{tikzpicture}
  \node (img)  {\includegraphics[width=8.1cm]{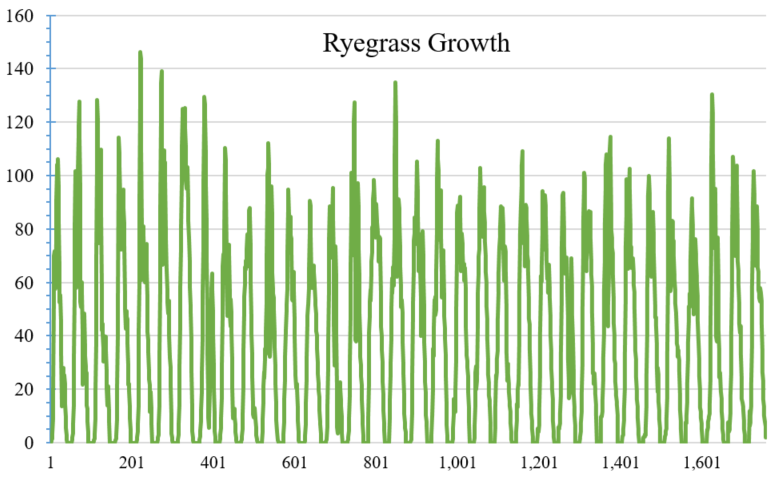}};
  \node[below=of img, node distance=0cm, yshift=1cm,] {Week number in the dataset};
  \node[left=of img, node distance=0cm, rotate=90, anchor=center,yshift=-0.8cm,] {Height (cm)};
\end{tikzpicture}
\caption{Weekly Grass Height at Moorepark From 1982-2013}
\label{fig:grass_height_weekly.png}
\end{figure}

\begin{figure}
\scriptsize
\centering
\begin{tikzpicture}
  \node (img)  {\includegraphics[width=8cm]{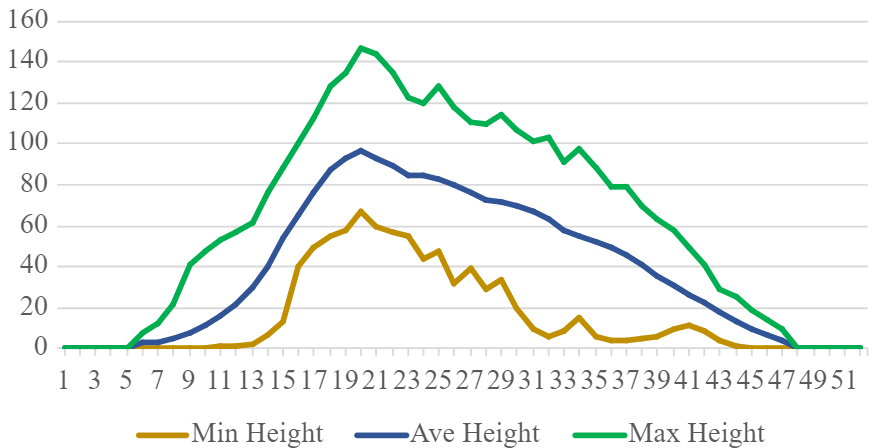}};
  \node[below=of img, node distance=0cm,   yshift=1cm,] {Week number in a year};
  \node[left=of img, node distance=0cm, rotate=90, anchor=center, xshift=0.5cm,  yshift=-0.8cm,] {Height (cm)};
\end{tikzpicture}
\caption{Weekly Height of Ryegrass}
\label{fig:grass_height_weekly.png}
\end{figure}

%


\section{Time Series Forecasting Models}\label{sec:models}
    
\subsection{AutoRegressive Integrated Moving Averge}

AutoRegressive Integrated Moving Average (ARIMA) represents a conventional statistical technique for time series forecasting \citep{KOTU2019395}. ARIMA integrates both the AutoRegressive (AR) process and the Moving Average (MA) process, making it a generalized form of ARMA. The AR process models temporal dependencies between an observation and a set of lagged observations in a given sequence. The hyperparameter $p$ denotes the number of lagged observations in the AR process. The general form of the AR process is presented in Equation \ref{eq:ar}, where $c$ is a constant, $\phi_{i}$ represents autoregression coefficients, $y_{i}$ is the dependent variable at time $i$, and $\epsilon_{t}$ is the error term for the predicted period.
    
    \begin{equation}
        \label{eq:ar}
        y_t = c + \sum^{p}_{i=1}\phi_{i} y_{t-i} * +\epsilon_{t}
    \end{equation}

The MA process models the temporal dependencies of the current observation as a function of past error terms that are independent of each other. It is represented in Equation \ref{eq:ma}, where $\mu$ is the mean of $y_{t}$, $\theta_{i}$ are parameters applied to the past error terms, and $\epsilon_{i}$ is a white noise error term at time $i$.

    \begin{equation}
        \label{eq:ma}
        y_t = \mu + \epsilon_{t} + \sum^{q}_{i=0}\theta_{i} \epsilon_{t-i}
    \end{equation}
    
A linear combination of the AR and MA process produces an ARMA process of order $(p,q)$ which can be mathematically expressed using equation \ref{eq:arma}, with $\phi_{i} \neq 0$ and $\theta_{i} \neq 0$.
    
    \begin{equation}
        \label{eq:arma}
        y_t = c + \sum^{p}_{i=1}\phi_{i} y_{t-i} + \epsilon_{t}+ \sum^{q}_{i=0}\theta_{i} \epsilon_{t-i}
    \end{equation}
    
ARIMA extends the ARMA model by incorporating an Integrated component, which accounts for non-seasonal differences in the time series.

\subsection{Long Short Term Memory}

Long Short-Term Memory stands out as a specialized variant of Recurrent Neural Network designed to preserve information from sequential data. It leverages patterns in the data and utilizes feedback loops to formulate predictions \citep{NGO2024106558}. LSTM effectively addresses the challenge of vanishing gradients by managing the information flow through a gated system, comprising a \emph{forget gate}, an \emph{input gate}, and an \emph{output gate}.

The \emph{input gate} and \emph{output gate} play pivotal roles in regulating the influx of input and output into the long-term state of the network. The forget gate determines the fraction of the long-term cell state that should be discarded. The mathematical representation of the LSTM cell is encapsulated in equations \ref{eq:lstm1} through \ref{eq:lstm6}.

  \vspace{-4mm}
\begin{equation}
    \label{eq:lstm1}
    i_t = \sigma(W_{i}x_{t} + U_{i}h_{t-1} - b_{i})
\end{equation}

\vspace{-4mm}
\begin{equation}
    \label{eq:lstm2}
    f_t = \sigma(W_{f}x_{t} + U_{f}h_{t-1} + b_{f})
\end{equation}

\vspace{-4mm}
\begin{equation}
    \label{eq:lstm3}
    o_t = \sigma(W_{o}x_{t} + U_{o}h_{t-1} - b_{o})
\end{equation}

\vspace{-4mm}
\begin{equation}
    \label{eq:lstm4}
    g_t = \phi(W_{g}x_{t} + U_{g}h_{t-1} - b_{g})
\end{equation}

\vspace{-4mm}
\begin{equation}
    \label{eq:lstm5}
    c_t = f_{t} \odot c_{t-1} + i_t \odot g_t
\end{equation}

\vspace{-4mm}
\begin{equation}
    \label{eq:lstm6}
    y_t = o_{t} \odot \phi(g_{t})
\end{equation}
    
In these equations, $W_{i}$, $W_{f}$, $W_{o}$, and $W_{g}$ represent weight matrices; $U_{i}$, $U_{f}$, $U_{o}$, and $U_{g}$ denote recurrent matrices; $b_{i}$, $b_{f}$, $b_{o}$, and $b_{g}$ are the bias terms; $h_{t-1}$ signifies the output at time $t-1$; $c_{t-1}$ indicates the cell state at time $t-1$; $\sigma$ represents the sigmoid activation function; $\phi$ denotes the hyperbolic activation function; and $\odot$ symbolizes element-wise multiplication of two vectors.

\subsection{Gated Recurrent Unit}

Gated Recurrent Unit (GRU) networks \citep{CMBB14, TANG2023103094} share similarities with LSTM as both employ gating mechanisms for information processing and recurrent connections. The primary distinction lies in the fact that GRU utilizes two gates: an update gate, $z_{t}$, and a reset gate, $r_{t}$, whereas LSTM employs three gates. Consequently, this simplification in the GRU network reduces the overall complexity by diminishing the number of parameters. The mathematical description of the GRU cell can be expressed through equations \ref{eq:gru1} to \ref{eq:gru4}.

\vspace{-4mm}
\begin{equation}
\label{eq:gru1}
    z_t = \sigma(W_{z}x_{t} + U_{z}h_{t-1} - b_{z})
\end{equation}

\vspace{-4mm}
\begin{equation}
\label{eq:gru2}
    r_t = \sigma(W_{r}x_{t} + U_{r}h_{t-1} + b_{r})
\end{equation}

\vspace{-4mm}
\begin{equation}
\label{eq:gru3}
    g_t = \phi(W_{g}x_{t} + U_{g}(r_{t} \odot h_{t-1}) + b_{g})
\end{equation}

\vspace{-4mm}
\begin{equation}
\label{eq:gru4}
    h_t = z_{t} \odot h_{t-1} + (1-z_{t}) \odot g_{t}
\end{equation}

Here, $W_{z}$, $W_{r}$, and $W_{g}$ serve as weight matrices; $U_{z}$, $U_{r}$, and $U_{g}$ act as recurrent matrices; and $b_{z}$, $b_{r}$, and $b_{g}$ are the associated bias terms. The variable $h_{t-1}$ represents the output at time $t-1$, with $h_{t}$ denoting the output at time $t$. The activation functions are also defined, with $\sigma$ representing a sigmoid activation and $\phi$ denoting a hyperbolic activation. Furthermore, the element-wise multiplication of two vectors is symbolized by $\odot$.

\subsection{Multi-Layer Perceptron}
    
A Multi-Layer Perceptron (MLP) consists of an input layer, hidden layers, and an output layer. The key distinction in the architectural composition of an MLP compared to a single-layer perceptron is the incorporation of hidden layers, classifying MLP as universal function approximators \citep{THSMAT21}. The "depth" of the MLP is determined by the number of hidden layers in the network, while the size of each hidden layer is defined by the number of neurons it contains. The connections between neurons in consecutive layers are modeled by weights, which are trainable parameters obtained through minimizing a cost function. In time series forecasting problems, Mean Squared Error is frequently employed as the loss function, as depicted in equation \ref{eq:mlp1}, where $C$ signifies the cost function, $y$ is the vector of true labels, $o$ is the vector of predictions, and $N$ denotes the number of samples in the training set.

\vspace{-4mm}
\begin{equation}
\label{eq:mlp1}
    C(y,o) = \frac{1}{N} \sum_{i=1}^{N}(y_{i} - o_{i})^2
\end{equation}

The gradients of the cost function are computed through backpropagation, an automated differentiation algorithm. To minimize the cost function, gradient descent is employed for optimization. Once the weights are determined, the output for a hidden layer in an MLP is computed using equations \ref{eq:mlp2} and \ref{eq:mlp3}.

\vspace{-4mm}
\begin{equation}
\label{eq:mlp2}
    a_l = W_{l}^{T}h_{l-1} + b_{l}
\end{equation}

\vspace{-4mm}
\begin{equation}
\label{eq:mlp3}
    h_l = \phi(a_{l})
\end{equation}

In these equations, $W_{l}^{T}$ represents a parameter matrix containing the weights at hidden layer $l$; $h_{l-1}$ denotes the input from the previous layer; $b_{l}$ is the bias term for the current layer; $a_{l}$ signifies the intermediate linear output; and $\phi$ denotes the activation function of the hidden layer.

\subsection{Temporal Convolutional Networks}

Temporal Convolutional Networks (TCN) are sequence models \citep{BKK18} and have found application in diverse domains such as speech detection \citep{WS21}, time series forecasting \citep{LCLR2020, LKR21}, anomaly detection in time series \citep{HZ19}, and action segmentation \citep{LFVRH17}.

While other sequence-based models like LSTM and GRU rely on recurrent connections for generating forecasts in a given time series, TCN adopts a distinct approach that harnesses the capabilities of the One-Dimensional Fully Convolutional Network (1D FCN) to generate an output sequence of the same length as the corresponding input sequence. TCN expands upon the 1D-FCN by incorporating causal convolutions instead of the standard convolutions found in the 1D-FCN. The key distinction lies in the fact that causal convolutions, when obtaining the output at time $t$, consider only elements from $t$ and past values from the preceding layers. This implies that, for a convolutional kernel of size $k$, zero padding of $k-1$ elements is asymmetrically applied at the beginning of the sequence.

\vspace{-2mm}
\begin{figure}[H]
    \centering
    \includegraphics[width=8cm]{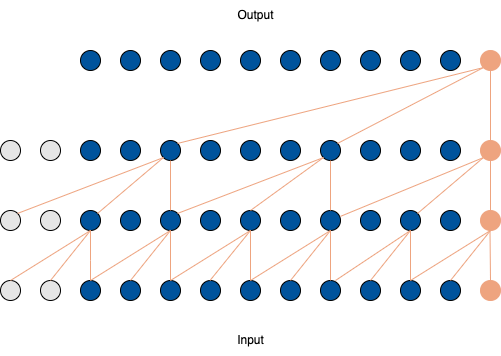}
    \caption{TCN model with 2 layers, a kernel size of 2 and dilations of [1, 2, 4]}
    \label{fig:avg_nn_layers.png}
\end{figure}
\vspace{-2mm}

The primary limitation of causal convolutions lies in the necessity for exceptionally large filters when modeling sequences with extended-range dependencies. This challenge can be addressed by introducing dilation to the convolutional filters, thereby expanding the receptive field and enabling the network to capture longer-range dependencies present in the input sequence. The dilated convolution operation $F$ on element $s$ of the sequence is illustrated in equation \ref{eq:tcn1}, where $f: {0,...,k -1}$ $\rightarrow$ $\mathbb{R}$ represents the convolution filter, $x$ is the input sequence, $k$ denotes the filter size, and $d$ is the dilation factor.

\vspace{-2mm}
\begin{equation}
\label{eq:tcn1}
    F(s) = \sum_{i=0}^{k-1}f(i) \cdot x_{s-d \cdot .i}
\end{equation}

The dilation factor experiences exponential growth as the network deepens, while the convolution kernel remains constant. This expansion of receptive fields enables the model to encompass all past values in the input sequence, as depicted in Figure \ref{fig:avg_nn_layers.png}.

\vspace{-2mm}
\begin{figure}[H]
    \centering
    \includegraphics[width=7cm]{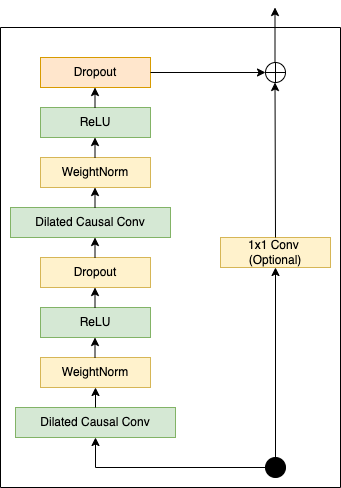}
    \caption{TCN Residual Block}
    \label{fig:resblock.png}
\end{figure}
\vspace{-2mm}

Residual connections play a crucial role in mitigating the issue of vanishing gradients in deep TCN models with numerous layers as Figure \ref{fig:resblock.png}. Our implementation is depicted in Equation \ref{eq:tcn2}, where $\theta$ represents the activation function, $s_{in}$ is the input sequence, $F$ stands for the identity mapping function that transforms the input sequence, and $O$ signifies the output.
    
\begin{equation}
\label{eq:tcn2}
    O = \theta (s_{in} + F(s_{in}))
\end{equation}

 
\section{Experimental Setup} \label{sec:exper}

\subsection{Dataset Processing}
\label{subsec:dataset_process}

To train and evaluate the supervised models, the dataset was divided into three subsets following a 60-20-20 split. The initial subset constituted the training set, encompassing 60\% of the entire dataset. The second subset served as the validation set, comprising 20\% of the dataset and utilized for hyperparameter tuning. The final 20\% subset was designated as the test set, employed for model evaluation.

Following the dataset split, the training set underwent MinMax Normalization to scale values within a [0, 1] range, a practice known to enhance convergence and performance in Neural Network models. The normalization parameters derived from the training set were then applied to transform the validation and test subsets. Each subset was further converted from a univariate sequence to a multivariate sequence, allowing the Neural Networks model to capture temporal dependencies among different lags of the time series and the current timestep's value. In this context, the term "lags" pertains to the length of the input sequence.

\subsection{Evaluation Metrics}
We employed the Root Mean Squared Error (RMSE) and the Mean Absolute Error (MAE) to assess the models' performance on the test set. RMSE is the square root of the mean of the squared errors as Equation \ref{eq:rmse}, a widely adopted metric recognized for its excellence in evaluating numerical prediction models \citep{ngo2025enhancing}.

\begin{equation}
\label{eq:rmse}
    RMSE = \sqrt{\frac{1}{N} \sum_{i=1}^{N}(y_{i} - x_{i})^2}
\end{equation}

\noindent
where $x_i$ represents observations, $y_i$ stands for predicted values of a variable, and N is the total number of observations. RMSE is a valuable accuracy metric, suitable for comparing forecasting errors across various models or configurations for a specific variable. However, it is scale-dependent, limiting its applicability for direct inter-variable comparisons.

MAE serves as a metric for the errors between paired observations representing the same phenomenon \citep{CHANG2024114155}, calculated as the sum of absolute errors divided by the size of the observation set, as illustrated in Equation \ref{eq:mae}. Here, $x_i$ denotes observations, $y_i$ represents predicted values of a variable, and N corresponds to the total number of observations.

\begin{equation}
\label{eq:mae}
    MAE = \frac{1}{N} \sum_{i=1}^{N}|y_{i} - x_{i}|
\end{equation}

\subsection{Hyperparameter Tuning}
\label{subsec:hyper}

The hyperparameters of the models assessed in this study underwent tuning through grid search on the validation set. For ARIMA, the tunable hyperparameter combinations involved the Autoregressive (AR) process $p$, with terms 1, 2, and 4, while the order of differencing included terms 1, 2, and 3. The Moving Average (MA) process $q$ encompassed terms 1, 2, and 4. This resulted in a total of 64 ARIMA$(p,d,q)$ models being evaluated, with the best-performing ARIMA model identified as ARIMA$(2,1,2)$.

The hyperparameter configurations considered for LSTM, GRU, and MLP models encompassed diverse settings. These included variations in the hidden layer size, ranging from a single layer with 5 or 10 neurons to multiple layers (2 or 3) with different neuron counts. Additionally, batch sizes of 32 and 64 units were explored. The input sequence length was examined across values of 2, 3, and 4, and a fixed number of epochs set at 50. The activation function for the hidden layers in all three models was Rectified Linear Units (ReLu), while the output layer employed a linear activation function. In total, this exhaustive exploration resulted in the evaluation of 108 distinct combinations of hyperparameters for LSTM, GRU, and MLP in the study.

For TCN, a comprehensive exploration of tunable hyperparameter combinations was conducted. This involved variations in the hidden layer size, ranging from 1 to 3, filters with options of 16, 32, and 64, convolutional kernel sizes of 2, 3, and 4, blocks spanning 2, 3, and 4, and dilations encompassing [1, 2, 4, 8, 16] and [1, 3, 6, 12, 24]. The input sequence length was examined across values of 2, 3, and 4, with a fixed number of epochs set at 30. The activation function for the hidden layers was ReLu, while the output layer employed a linear activation function. This thorough investigation resulted in the evaluation of a total of 486 TCN models in this study.

\section{Results} \label{sec:results}

\subsection{Effect of Network Layers and Input Sequence Length}\label{AA}

The table \ref{tab:layers_acc} depicts the average RMSE scores for neural network models employing varying numbers of layers. Notably, all models, except for TCN, exhibited a preference for a 2-layer architecture on average, whereas TCN favored a 1-layer model architecture. It is noteworthy that a consistent decline in performance was observed across all models when transitioning from a 2-layer to a 3-layer architecture. LSTM exhibited comparable performance between the single-layer architecture and the 2-layer architectures, a distinctive pattern compared to the other models examined in this study. Notably, TCN demonstrated the widest range (447) between the 2-layer and 3-layer configurations, achieving the best overall performance, whereas GRU and MLP had narrower ranges of 1.81 and 2.14, respectively. This emphasizes the critical importance of thoughtfully selecting layer configurations, especially when employing TCN to model grass growth in time series.

\begin{table}[H]
\caption{Average RMSE scores for different layers}
\vspace{-5mm}
\begin{center}
\begin{tabular}{c|c|c|c|c}
    \hline
    \hline
    \textbf{Layer} & \textbf{LSTM} & \textbf{GRU} & \textbf{MLP} & \textbf{TCN}\\
    \hline
    1 & 11.33 & 10.65 & 7.46 & \textbf{4.63}  \\
    \hline
    2 & \textbf{10.92} & \textbf{6.55} & \textbf{5.14} & 15.77 \\
    \hline
    3 & 15.62 & 8.36 & 7.28 & 462.77  \\
    \hline
    \hline
\end{tabular}
\label{tab:layers_acc}
\end{center}
\end{table}

Table \ref{tab:seq_len} presents the average RMSE scores for Neural Network models with varying input sequence lengths. On average, the optimal input sequence length for LSTM, MLP and TCN was the preceding two weeks, while GRU performed best when utilizing the preceding three weeks for predicting future grass growth. Notably, both MLP and TCN exhibited consistent degradation in performance as the sequence length increased, whereas the LSTM and GRU experienced slight performance boosts in certain cases with an extended sequence length.
    
\begin{table}[H]
\caption{Average RMSE scores for different sequence lengths}
\vspace{-5mm}
\begin{center}
\begin{tabular}{c|c|c|c|c}
    \hline
    \hline
    \textbf{Length} & \textbf{LSTM} & \textbf{GRU} & \textbf{MLP} & \textbf{TCN}\\
    \hline
    2 & \textbf{14.89} & 14.48 & \textbf{7.60} & \textbf{72.24}  \\
    \hline
    3 & 17.83 & \textbf{8.35} & 9.12 & 222.03 \\
    \hline
    4 & 15.94 & 10.78 & 10.32 & 728.33  \\
    \hline
    \hline
\end{tabular}
\label{tab:seq_len}
\end{center}
\end{table}

\subsection{The Best Performance}



\begin{table}[htpb]
\caption{The Best Performances}
\vspace{-4mm}
\begin{center}
\begin{tabular}{c|c|c|c|c|c}
    \hline
    \hline
    \textbf{Model} & \textbf{RMSE}& \textbf{MAE} & \textbf{RT(s)} & \textbf{Layer} & \textbf{Length}\\ \hline
    ARIMA & 5.31 & 4.07 & \textbf{1.08} & N/A & N/A \\ \hline
    LSTM & 5.94 & 5.39 & 3.69 & 1 & 3 \\ \hline
    GRU & 3.75 & 4.48 & 5.55 & 1 & 3 \\ \hline
    MLP & 3.64 & 4.15 & 1.56 & 1 & 2 \\ \hline
    TCN & \textbf{2.74} & \textbf{3.46} & 69.51 & 1 & 2 \\ \hline
    \hline
\end{tabular}
\label{tab:best_results}
\end{center}
\end{table}

Table \ref{tab:best_results} displays the optimal MAE and RMSE performances achieved by all models on the grass dataset. The most successful model in this experimental study is the TCN, attaining an RMSE of 2.74 and an MAE of 3.46. Conversely, the least effective model originates from the LSTM, registering an RMSE of 5.94 and an MAE of 5.39. Compared to the baseline ARIMA model, GRU, MLP, and TCD exhibit better performance in terms of RMSE. However, when considering MAE, only TCN shows improved results. Comparatively, the TCN outperforms ARIMA, LSTM, GRU, and MLP by 48.4\%, 53.9\%, 26.9\%, and 24.7\% in terms of RMSE, respectively. In MAE, the superiority of TCN over ARIMA, LSTM, GRU, and MLP is evident, surpassing them by 15.0\%, 35.8\%, 22.8\%, and 16.6\%, respectively. 

The optimal configurations for the models are a single layer and a sequence length of 3 for both LSTM and GRU. Meanwhile, MLP and TCN perform optimally with a single layer and a sequence length of 2. Additionally, the number of nodes and batch size are set to 10 and 32, respectively, for LSTM, GRU, and MLP. For TCN, the optimal settings include 64 filters, a kernel size of 4, and a block size of 3. Furthermore, TCN employs a dilation of [1, 3, 6, 12, 24].

In terms of runtimes, the DL models investigated in this study differ from simpler ML models. This discrepancy arises primarily from their architectural complexity and the necessity for a larger number of trainable parameters, contributing to extended training times. The baseline ARIMA model boasts the quickest runtime at 1.08 seconds. Following closely, the second-fastest model is MLP, completing in 1.56 seconds. In contrast, LSTM, GRU, and TCN exhibit longer runtimes, clocking in at 3.69, 5.55, and 69.51 seconds, respectively.

In DL models, a noticeable distinction exists between TCN and other models, primarily stemming from the TCN model's requirement of a higher number of trainable parameters, leading to significantly longer training times. In contrast, the simplicity of the MLP model allows it to consistently operate at lower computational times for any hyperparameter combination. The LSTM and GRU exhibit very similar results in terms of training times, a phenomenon explained by the similarity in their model architectures. Notably, substantial increases in running times for TCN occur when the number of layers in the network increases, as opposed to more modest increases when the length of the input sequence increases. This trend is unique to TCN, unlike GRU, LSTM, and MLP, which do not exhibit such pronounced increases in running times with model complexity.

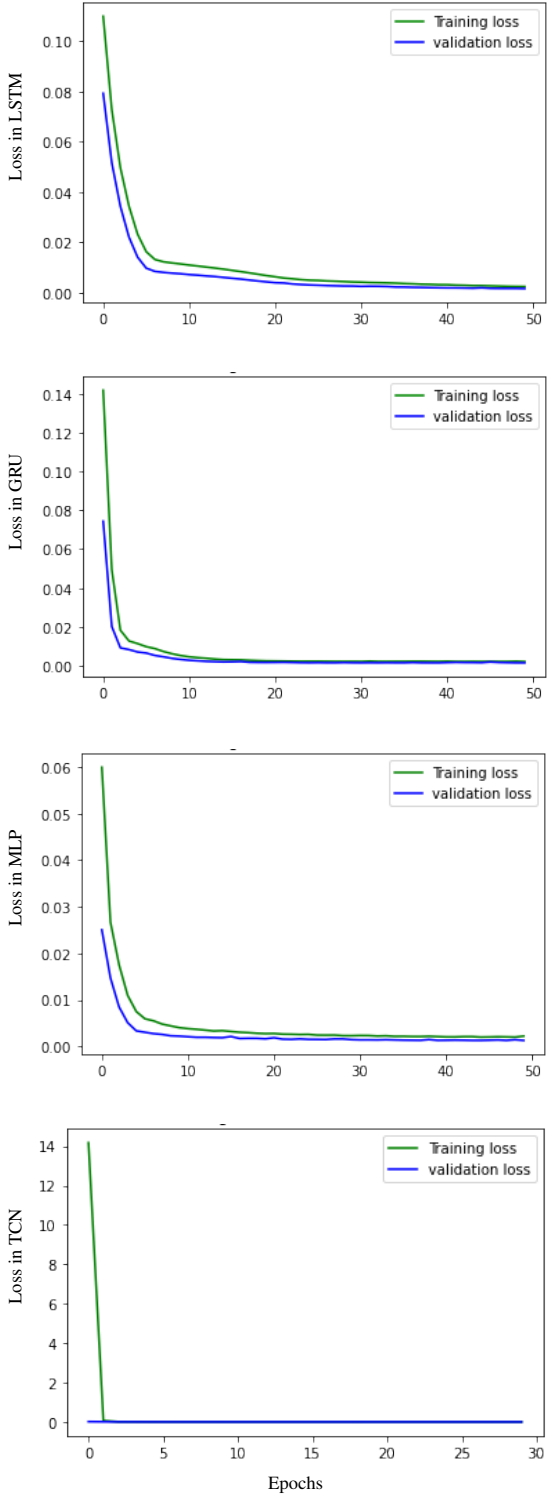
\begin{figure}[H]
\scriptsize
\centering
\begin{tikzpicture}
  \node (lstm)  {\includegraphics[width=6.8cm]{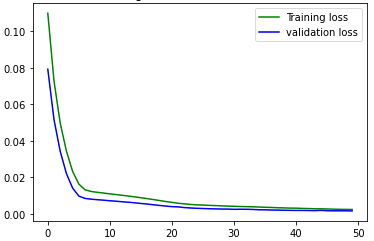}};
  \node[left=of lstm, node distance=0cm, rotate=90, anchor=center, xshift=0.5cm,  yshift=-0.8cm,] {Loss in LSTM};

  \node (gru)[below=of lstm, yshift= 0.6cm]  {\includegraphics[width=6.8cm]{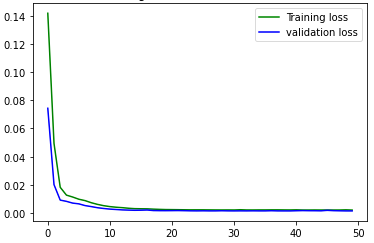}};
  \node[left=of gru, node distance=0cm, rotate=90, anchor=center, xshift=0.5cm,  yshift=-0.8cm,] {Loss in GRU};

  \node (mlp)[below=of gru, yshift= 0.6cm]  {\includegraphics[width=6.8cm]{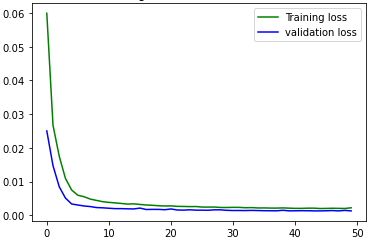}};
  \node[left=of mlp, node distance=0cm, rotate=90, anchor=center, xshift=0.5cm,  yshift=-0.8cm,] {Loss in MLP};

  \node (tcn)[below=of mlp, yshift= 0.6cm]  {\includegraphics[width=6.8cm]{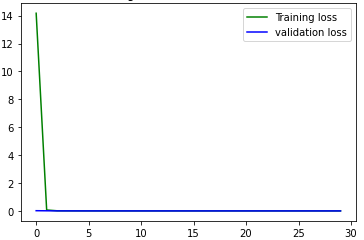}};
  \node[left=of tcn, node distance=0cm, rotate=90, anchor=center, xshift=0.5cm,  yshift=-0.8cm,] {Loss in TCN};
  \node[below=of tcn, node distance=0cm,   yshift=1cm,] {Epochs};
  
\end{tikzpicture}
\caption{Showing the training and validation loss curves of the experimental models}
\label{fig:val_loss}
\end{figure}

The training and validation loss curves, as illustrated in Figure \ref{fig:val_loss}, serve as insightful indicators of the models' learning dynamics. The consistent decrease in both curves not only suggests effective learning but also serves as a safeguard against overfitting, emphasizing the models' ability to generalize well. It is particularly noteworthy that TCN exhibits a remarkable level of uniformity in both its training and validation curves, surpassing the performance of the other models.

Moreover, the lower validation loss observed in TCN is a crucial metric highlighting its prowess in generalizing to previously unseen data. This lower validation loss underscores TCN's capacity to extrapolate learned patterns beyond the training set, making it a promising candidate for real-world applications where robust generalization is paramount. Consequently, the combined evidence from the uniformity in training and validation curves, along with the superior performance in validation loss, positions TCN as a standout model in the context of effective learning and generalization. 

\subsection{Discussions}

Predicting future grass growth holds the potential to enhance resource utilization and optimize grassland management for farmers, thereby minimizing the reliance on external feed purchases. While decision support tools like PastureBase-Ireland\footnote{https://pasturebase.teagasc.ie/V2/login.aspx} have been developed in the past, many lack forecasting capabilities, limiting their overall efficiency. The importance of incorporating forecasting into these tools is increasingly evident, especially considering the anticipated climate changes that are expected to introduce more variability within and between years. This heightened variability underscores the necessity for farmers to be more reactive and flexible in their approaches, making the ability to forecast grass growth a crucial element for sustainable and adaptive grassland management practices.

The utilization of statistical models has emerged as a promising approach for modeling biological processes, gaining significant traction across diverse domains like epidemiology\citep{NKAS23}, ecology\citep{ANCEGKLBMF21}, and agronomy\citep{ZS21}. This study focuses on exploring time series forecasting, a crucial subset of these statistical modeling techniques, as potential surrogates\citep{GAPGGM23}. These models serve as valuable tools for approximating the intricate and interrelated biological interactions essential for understanding the dynamics of grass growth.

The time series depicting grass growth exhibits a consistent trend marked by undulating peaks and troughs that align with climatic seasons, as illustrated in Figure \ref{fig:grass_height_weekly.png}. Transforming this time series into a stationary form allows us to uncover valuable properties, laying the groundwork for forecasting future values. In our investigation, ARIMA served as the baseline model, initiating a comparison with deep learning algorithms, specifically LSTM, for time series forecasting.

Contrary to expectations, ARIMA has proven to outperform LSTM and perform comparably to GRU in this study. One potential explanation for this unexpected result may be found in the granularity of the time scales considered. As observed in a similar time series forecasting task detailed in \citep{ZSCWWL22}, ARIMA exhibited superior performance over LSTM for monthly and weekly forecasts, although not for daily forecasts. The granularity of time scales often directly impacts the size of the input dataset. Therefore, the observed efficacy of ARIMA over LSTM in our study could be attributed to the relative size of our input dataset. This underscores the importance of carefully considering the specific characteristics of the data and the time scales involved when selecting an appropriate forecasting model.

Recently, TCN has gained popularity for their effectiveness in numerous time series forecasting tasks. In our study, TCN demonstrated superior performance compared to ARIMA and other deep learning models. However, it's important to note that the heightened complexity of TCN demands thorough experimentation on the input dataset before practical implementation. The sensitivity of the model to its parameters underscores the necessity for careful tuning to ensure optimal performance. This finding emphasizes the potential of TCN as powerful tools in time series forecasting while highlighting the importance of meticulous parameter adjustment to harness their full capabilities.

\section{Conclusion and Future Work} \label{sec:conclusion}

This paper presents a comparative analysis of ARIMA, GRU, LSTM, MLP, and TCN in forecasting grass growth time series. TCN exhibited superior performance, outperforming all other evaluated models. Contrarily, the recurrent models failed to surpass the baseline and were outperformed even by simpler architectures like MLP. The optimal sequence window length was found to be model-dependent, and the advantages of increasing model complexity in the case of deep learning models translated into only marginal performance gains, with minor exceptions for GRU and LSTM.

In future research, we aim to enhance our current results by incorporating climatic parameters as features. This introduces a multivariate aspect to the task, prompting the need for new approaches tailored to this complexity. Additionally, a relevant avenue for future work involves leveraging domain expertise to engineer new features, thereby improving the models' performance. Furthermore, we will explore the integration of ontologies \citep{cao2012semantic}, knowledge graphs \citep{ngol2021semantic,Ngo.2025}, time series prediction \citep{Bahrpeyma2025}  and data warehouse frameworks \citep{ngo2020data} to enhance both classification performance and computational efficiency.


\section*{Funding Acknowledgement} This work was supported by Science Foundation Ireland through the Insight Centre for Data Analytics (SFI/12/RC/2289\_P2) and Vistamilk (SFI/16/RC/3835).

\bibliographystyle{elsarticle-harv} 
\bibliography{reference.bib}

\begin{thebibliography}{48}
\expandafter\ifx\csname natexlab\endcsname\relax\def\natexlab#1{#1}\fi
\providecommand{\url}[1]{\texttt{#1}}
\providecommand{\href}[2]{#2}
\providecommand{\path}[1]{#1}
\providecommand{\DOIprefix}{doi:}
\providecommand{\ArXivprefix}{arXiv:}
\providecommand{\URLprefix}{URL: }
\providecommand{\Pubmedprefix}{pmid:}
\providecommand{\doi}[1]{\href{http://dx.doi.org/#1}{\path{#1}}}
\providecommand{\Pubmed}[1]{\href{pmid:#1}{\path{#1}}}
\providecommand{\bibinfo}[2]{#2}
\ifx\xfnm\relax \def\xfnm[#1]{\unskip,\space#1}\fi
\bibitem[{Ali et~al.(2016)Ali, Cawkwell, Dwyer, Barrett and Green}]{ACDBG16}
\bibinfo{author}{Ali, I.}, \bibinfo{author}{Cawkwell, F.}, \bibinfo{author}{Dwyer, E.}, \bibinfo{author}{Barrett, B.}, \bibinfo{author}{Green, S.}, \bibinfo{year}{2016}.
\newblock \bibinfo{title}{Satellite remote sensing of grasslands: From observation to management}.
\newblock \bibinfo{journal}{Journal of Plant Ecology} \bibinfo{volume}{9}, \bibinfo{pages}{649–671}.
\newblock \DOIprefix\doi{10.1093/jpe/rtw005}.
\bibitem[{Auger‐Méthé et~al.(2021)Auger‐Méthé, Newman, Cole, Empacher, Gryba, King, Leos‐Barajas, Mills~Flemming, Nielsen, Petris and et~al.}]{ANCEGKLBMF21}
\bibinfo{author}{Auger‐Méthé, M.}, \bibinfo{author}{Newman, K.}, \bibinfo{author}{Cole, D.}, \bibinfo{author}{Empacher, F.}, \bibinfo{author}{Gryba, R.}, \bibinfo{author}{King, A.A.}, \bibinfo{author}{Leos‐Barajas, V.}, \bibinfo{author}{Mills~Flemming, J.}, \bibinfo{author}{Nielsen, A.}, \bibinfo{author}{Petris, G.}, \bibinfo{author}{et~al.}, \bibinfo{year}{2021}.
\newblock \bibinfo{title}{A guide to state–space modeling of ecological time series}.
\newblock \bibinfo{journal}{Ecological Monographs} \bibinfo{volume}{91}.
\newblock \DOIprefix\doi{10.1002/ecm.1470}.
\bibitem[{Bahrpeyma et~al.(2025)Bahrpeyma, Ngo, Roantree and McCarren}]{Bahrpeyma2025}
\bibinfo{author}{Bahrpeyma, F.}, \bibinfo{author}{Ngo, V.M.}, \bibinfo{author}{Roantree, M.}, \bibinfo{author}{McCarren, A.}, \bibinfo{year}{2025}.
\newblock \bibinfo{title}{A meta-learner approach to multistep-ahead time series prediction}.
\newblock \bibinfo{journal}{International Journal of Data Science and Analytics} \bibinfo{volume}{20}, \bibinfo{pages}{2291--2303}.
\newblock \DOIprefix\doi{10.1007/s41060-024-00599-6}.
\bibitem[{Bai et~al.(2018)Bai, Kolter and Koltun}]{BKK18}
\bibinfo{author}{Bai, S.B.}, \bibinfo{author}{Kolter, J.Z.}, \bibinfo{author}{Koltun, V.}, \bibinfo{year}{2018}.
\newblock \bibinfo{title}{An empirical evaluation of generic convolutional and recurrent networks for sequence modeling}.
\newblock \bibinfo{journal}{CoRR} \bibinfo{volume}{abs/1803.01271}.
\newblock \URLprefix \url{http://arxiv.org/abs/1803.01271}.
\bibitem[{Beck et~al.(2018)Beck, Zimmermann, McVicar, Vergopolan, Berg and Wood}]{BZMV18}
\bibinfo{author}{Beck, H.E.}, \bibinfo{author}{Zimmermann, N.E.}, \bibinfo{author}{McVicar, T.R.}, \bibinfo{author}{Vergopolan, N.}, \bibinfo{author}{Berg, A.}, \bibinfo{author}{Wood, E.F.}, \bibinfo{year}{2018}.
\newblock \bibinfo{title}{Present and future köppen-geiger climate classification maps at 1-km resolution}.
\newblock \bibinfo{journal}{Scientific Data} \bibinfo{volume}{5}.
\newblock \DOIprefix\doi{10.1038/sdata.2018.214}.
\bibitem[{Benedict et~al.(2023)Benedict, Boyte, Dahal, Shrestha, Parajuli and Megard}]{Benedict:2023}
\bibinfo{author}{Benedict, T.D.}, \bibinfo{author}{Boyte, S.P.}, \bibinfo{author}{Dahal, D.}, \bibinfo{author}{Shrestha, D.}, \bibinfo{author}{Parajuli, S.}, \bibinfo{author}{Megard, L.J.}, \bibinfo{year}{2023}.
\newblock \bibinfo{title}{Extracting exotic annual grass phenology and climate relations in western u.s. rangeland ecoregions}.
\newblock \bibinfo{journal}{Biological Invasions} \bibinfo{volume}{25}, \bibinfo{pages}{2023–2041}.
\newblock \DOIprefix\doi{10.1007/s10530-023-03021-7}.
\bibitem[{Cao and Ngo(2012)}]{cao2012semantic}
\bibinfo{author}{Cao, T.H.}, \bibinfo{author}{Ngo, V.M.}, \bibinfo{year}{2012}.
\newblock \bibinfo{title}{Semantic search by latent ontological features}.
\newblock \bibinfo{journal}{New Generation Computing} \bibinfo{volume}{30}, \bibinfo{pages}{53--71}.
\bibitem[{Chang et~al.(2024)Chang, Lu, Guo, Zhou and Xiu}]{CHANG2024114155}
\bibinfo{author}{Chang, L.}, \bibinfo{author}{Lu, Q.}, \bibinfo{author}{Guo, Y.}, \bibinfo{author}{Zhou, B.}, \bibinfo{author}{Xiu, G.}, \bibinfo{year}{2024}.
\newblock \bibinfo{title}{Error correction algorithm for grating moiré fringes based on qm-ann}.
\newblock \bibinfo{journal}{Measurement} \bibinfo{volume}{226}, \bibinfo{pages}{114155}.
\newblock \DOIprefix\doi{10.1016/j.measurement.2024.114155}.
\bibitem[{Cho et~al.(2014)Cho, van Merrienboer, Bahdanau and Bengio}]{CMBB14}
\bibinfo{author}{Cho, K.}, \bibinfo{author}{van Merrienboer, B.}, \bibinfo{author}{Bahdanau, D.}, \bibinfo{author}{Bengio, Y.}, \bibinfo{year}{2014}.
\newblock \bibinfo{title}{On the properties of neural machine translation: Encoder–decoder approaches}.
\newblock \bibinfo{journal}{Proceedings of SSST-8, Eighth Workshop on Syntax, Semantics and Structure in Statistical Translation} \DOIprefix\doi{10.3115/v1/w14-4012}.
\bibitem[{Defalque et~al.(2024)Defalque, Santos, Bungenstab, Echeverria, Dias and Defalque}]{Defalque:2024}
\bibinfo{author}{Defalque, G.}, \bibinfo{author}{Santos, R.}, \bibinfo{author}{Bungenstab, D.}, \bibinfo{author}{Echeverria, D.}, \bibinfo{author}{Dias, A.}, \bibinfo{author}{Defalque, C.}, \bibinfo{year}{2024}.
\newblock \bibinfo{title}{Machine learning models for dry matter and biomass estimates on cattle grazing systems}.
\newblock \bibinfo{journal}{Computers and Electronics in Agriculture} \bibinfo{volume}{216}, \bibinfo{pages}{108520}.
\newblock \DOIprefix\doi{10.1016/j.compag.2023.108520}.
\bibitem[{Fitzgerald(2019)}]{Fitzgerald:2019}
\bibinfo{author}{Fitzgerald, C.}, \bibinfo{year}{2019}.
\newblock \bibinfo{title}{Dairy in the irish economy - growing sustainably}.
\newblock \URLprefix \url{https://www.teagasc.ie/media/website/publications/2019/Dairy-in-the-Irish-economy.pdf}.
\bibitem[{Gherman et~al.(2023)Gherman, Abdallah, Pang, Gorochowski, Grierson and Marucci}]{GAPGGM23}
\bibinfo{author}{Gherman, I.M.}, \bibinfo{author}{Abdallah, Z.S.}, \bibinfo{author}{Pang, W.}, \bibinfo{author}{Gorochowski, T.E.}, \bibinfo{author}{Grierson, C.S.}, \bibinfo{author}{Marucci, L.}, \bibinfo{year}{2023}.
\newblock \bibinfo{title}{Bridging the gap between mechanistic biological models and machine learning surrogates}.
\newblock \bibinfo{journal}{PLOS Computational Biology} \bibinfo{volume}{19}.
\newblock \DOIprefix\doi{10.1371/journal.pcbi.1010988}.
\bibitem[{Hanrahan et~al.(2017)Hanrahan, Geoghegan, O'Donovan, Griffith, Ruelle, Wallace and Shalloo}]{HGO17}
\bibinfo{author}{Hanrahan, L.}, \bibinfo{author}{Geoghegan, A.}, \bibinfo{author}{O'Donovan, M.}, \bibinfo{author}{Griffith, V.}, \bibinfo{author}{Ruelle, E.}, \bibinfo{author}{Wallace, M.}, \bibinfo{author}{Shalloo, L.}, \bibinfo{year}{2017}.
\newblock \bibinfo{title}{Pasturebase ireland: A grassland decision support system and national database}.
\newblock \bibinfo{journal}{Computers and Electronics in Agriculture} \bibinfo{volume}{136}, \bibinfo{pages}{193–201}.
\newblock \DOIprefix\doi{10.1016/j.compag.2017.01.029}.
\bibitem[{He and Zhao(2019)}]{HZ19}
\bibinfo{author}{He, Y.}, \bibinfo{author}{Zhao, J.}, \bibinfo{year}{2019}.
\newblock \bibinfo{title}{Temporal convolutional networks for anomaly detection in time series}.
\newblock \bibinfo{journal}{Journal of Physics: Conference Series} \bibinfo{volume}{1213}, \bibinfo{pages}{042050}.
\newblock \DOIprefix\doi{10.1088/1742-6596/1213/4/042050}.
\bibitem[{Holtgrave et~al.(2023)Holtgrave, Lobert, Erasmi, Röder and Kleinschmit}]{Holtgrave:2023}
\bibinfo{author}{Holtgrave, A.K.}, \bibinfo{author}{Lobert, F.}, \bibinfo{author}{Erasmi, S.}, \bibinfo{author}{Röder, N.}, \bibinfo{author}{Kleinschmit, B.}, \bibinfo{year}{2023}.
\newblock \bibinfo{title}{Grassland mowing event detection using combined optical, sar, and weather time series}.
\newblock \bibinfo{journal}{Remote Sensing of Environment} \bibinfo{volume}{295}, \bibinfo{pages}{113680}.
\newblock \DOIprefix\doi{10.1016/j.rse.2023.113680}.
\bibitem[{Irish-Food-Board(2022)}]{Irish-FB:2022}
\bibinfo{author}{Irish-Food-Board}, \bibinfo{year}{2022}.
\newblock \bibinfo{title}{Export performance and prospects report 2021-2022 in ireland's agri-food sector}.
\newblock \URLprefix \url{https://www.bordbia.ie/globalassets/performance-and-prospects/bord-bias-export-performance--prospects-2021---2022-pdf-report.pdf}.
\bibitem[{Kenny et~al.(2019)Kenny, Ruelle, Geoghegan, Shalloo, O’Leary, O’Donovan and Keane}]{KRGS19}
\bibinfo{author}{Kenny, E.M.}, \bibinfo{author}{Ruelle, E.}, \bibinfo{author}{Geoghegan, A.}, \bibinfo{author}{Shalloo, L.}, \bibinfo{author}{O’Leary, M.}, \bibinfo{author}{O’Donovan, M.}, \bibinfo{author}{Keane, M.T.}, \bibinfo{year}{2019}.
\newblock \bibinfo{title}{Predicting grass growth for sustainable dairy farming: A cbr system using bayesian case-exclusion and post-hoc, personalized explanation-by-example (xai)}.
\newblock \bibinfo{journal}{Case-Based Reasoning Research and Development} , \bibinfo{pages}{172–187}\DOIprefix\doi{10.1007/978-3-030-29249-2_12}.
\bibitem[{Kotu and Deshpande(2019)}]{KOTU2019395}
\bibinfo{author}{Kotu, V.}, \bibinfo{author}{Deshpande, B.}, \bibinfo{year}{2019}.
\newblock \bibinfo{title}{Chapter 12 - time series forecasting}, in: \bibinfo{editor}{Kotu, V.}, \bibinfo{editor}{Deshpande, B.} (Eds.), \bibinfo{booktitle}{Data Science (Second Edition)}. \bibinfo{edition}{second edition} ed.. \bibinfo{publisher}{Morgan Kaufmann}, pp. \bibinfo{pages}{395--445}.
\newblock \DOIprefix\doi{10.1016/B978-0-12-814761-0.00012-5}.
\bibitem[{Lara-Benítez et~al.(2020)Lara-Benítez, Carranza-García, Luna-Romera and Riquelme}]{LCLR2020}
\bibinfo{author}{Lara-Benítez, P.}, \bibinfo{author}{Carranza-García, M.}, \bibinfo{author}{Luna-Romera, J.M.}, \bibinfo{author}{Riquelme, J.C.}, \bibinfo{year}{2020}.
\newblock \bibinfo{title}{Temporal convolutional networks applied to energy-related time series forecasting}.
\newblock \bibinfo{journal}{Applied Sciences} \bibinfo{volume}{10}, \bibinfo{pages}{2322}.
\newblock \DOIprefix\doi{10.3390/app10072322}.
\bibitem[{Latham et~al.(2014)Latham, Cumani, Rosati and Bloise}]{Latham:2014}
\bibinfo{author}{Latham, J.}, \bibinfo{author}{Cumani, R.}, \bibinfo{author}{Rosati, I.}, \bibinfo{author}{Bloise, M.}, \bibinfo{year}{2014}.
\newblock \bibinfo{title}{Global land cover share (glc-share): Database beta-release version 1.0}.
\newblock \URLprefix \url{https://www.fao.org/uploads/media/glc-share-doc.pdf}.
\bibitem[{Lea et~al.(2017)Lea, Flynn, Vidal, Reiter and Hager}]{LFVRH17}
\bibinfo{author}{Lea, C.}, \bibinfo{author}{Flynn, M.D.}, \bibinfo{author}{Vidal, R.}, \bibinfo{author}{Reiter, A.}, \bibinfo{author}{Hager, G.D.}, \bibinfo{year}{2017}.
\newblock \bibinfo{title}{Temporal convolutional networks for action segmentation and detection}.
\newblock \bibinfo{journal}{2017 IEEE Conference on Computer Vision and Pattern Recognition (CVPR)} \DOIprefix\doi{10.1109/cvpr.2017.113}.
\bibitem[{Lin et~al.(2021)Lin, Koprinska and Rana}]{LKR21}
\bibinfo{author}{Lin, Y.}, \bibinfo{author}{Koprinska, I.}, \bibinfo{author}{Rana, M.}, \bibinfo{year}{2021}.
\newblock \bibinfo{title}{Temporal convolutional attention neural networks for time series forecasting}.
\newblock \bibinfo{journal}{2021 International Joint Conference on Neural Networks (IJCNN)} \DOIprefix\doi{10.1109/ijcnn52387.2021.9534351}.
\bibitem[{Lindblom et~al.(2017)Lindblom, Lundström, Ljung and Jonsson}]{Jessica:2017}
\bibinfo{author}{Lindblom, J.}, \bibinfo{author}{Lundström, C.}, \bibinfo{author}{Ljung, M.}, \bibinfo{author}{Jonsson, A.}, \bibinfo{year}{2017}.
\newblock \bibinfo{title}{Promoting sustainable intensification in precision agriculture: review of decision support systems development and strategies} \bibinfo{volume}{18}, \bibinfo{pages}{309–331}.
\newblock \DOIprefix\doi{10.1007/s11119-016-9491-4}.
\bibitem[{Lussem et~al.(2019)Lussem, Bolten, Menne, Gnyp, Schellberg and Bareth}]{Lussem:2019}
\bibinfo{author}{Lussem, U.}, \bibinfo{author}{Bolten, A.}, \bibinfo{author}{Menne, J.}, \bibinfo{author}{Gnyp, M.L.}, \bibinfo{author}{Schellberg, J.}, \bibinfo{author}{Bareth, G.}, \bibinfo{year}{2019}.
\newblock \bibinfo{title}{Estimating biomass in temperate grassland with high resolution canopy surface models from uav-based rgb images and vegetation indices}.
\newblock \bibinfo{journal}{Journal of Applied Remote Sensing} \bibinfo{volume}{13}, \bibinfo{pages}{034525}.
\newblock \DOIprefix\doi{10.1117/1.jrs.13.034525}.
\bibitem[{McHugh et~al.(2020)McHugh, Browne, Liu and Jordan}]{MBLJ20}
\bibinfo{author}{McHugh, O.}, \bibinfo{author}{Browne, F.}, \bibinfo{author}{Liu, J.}, \bibinfo{author}{Jordan, P.}, \bibinfo{year}{2020}.
\newblock \bibinfo{title}{A decision analytic framework and exploratory statistical case study analysis of grass growth in northern ireland}.
\newblock \bibinfo{journal}{Journal of Advances in Information Technology} , \bibinfo{pages}{15–20}\DOIprefix\doi{10.12720/jait.11.1.15-20}.
\bibitem[{Murphy et~al.(2021)Murphy, Murphy, O’Brien and O’Donovan}]{MMBD21}
\bibinfo{author}{Murphy, D.J.}, \bibinfo{author}{Murphy, M.D.}, \bibinfo{author}{O’Brien, B.}, \bibinfo{author}{O’Donovan, M.}, \bibinfo{year}{2021}.
\newblock \bibinfo{title}{A review of precision technologies for optimising pasture measurement on irish grassland}.
\newblock \bibinfo{journal}{Agriculture} \bibinfo{volume}{11}, \bibinfo{pages}{600}.
\newblock \DOIprefix\doi{10.3390/agriculture11070600}.
\bibitem[{Nagvanshi et~al.(2023)Nagvanshi, Kaur, Agarwal and Sharma}]{NKAS23}
\bibinfo{author}{Nagvanshi, S.S.}, \bibinfo{author}{Kaur, I.}, \bibinfo{author}{Agarwal, C.}, \bibinfo{author}{Sharma, A.}, \bibinfo{year}{2023}.
\newblock \bibinfo{title}{Nonstationary time series forecasting using optimized-evdhm-arima for covid-19}.
\newblock \bibinfo{journal}{Frontiers in Big Data} \bibinfo{volume}{6}.
\newblock \DOIprefix\doi{10.3389/fdata.2023.1081639}.
\bibitem[{Ngo et~al.(2025a)Ngo, Bolger, Goodwin, O'Sullivan, Cuong and Roantree}]{Ngo.2025}
\bibinfo{author}{Ngo, V.M.}, \bibinfo{author}{Bolger, E.}, \bibinfo{author}{Goodwin, S.}, \bibinfo{author}{O'Sullivan, J.}, \bibinfo{author}{Cuong, D.V.}, \bibinfo{author}{Roantree, M.}, \bibinfo{year}{2025}a.
\newblock \bibinfo{title}{A graph based raman spectral processing technique for exosome classification}, in: \bibinfo{editor}{Bellazzi, R.}, et~al. (Eds.), \bibinfo{booktitle}{Artificial Intelligence in Medicine}, \bibinfo{publisher}{Springer}. pp. \bibinfo{pages}{344--354}.
\newblock \DOIprefix\doi{10.1007/978-3-031-95838-0_34}.
\bibitem[{Ngo et~al.(2023)Ngo, Duong, Nguyen, Dang and Conlan}]{Ngo:2023}
\bibinfo{author}{Ngo, V.M.}, \bibinfo{author}{Duong, T.V.T.}, \bibinfo{author}{Nguyen, T.B.T.}, \bibinfo{author}{Dang, C.N.}, \bibinfo{author}{Conlan, O.}, \bibinfo{year}{2023}.
\newblock \bibinfo{title}{A big data smart agricultural system: recommending optimum fertilisers for crops}.
\newblock \bibinfo{journal}{International Journal of Information Technology} \bibinfo{volume}{15}, \bibinfo{pages}{249–265}.
\newblock \DOIprefix\doi{10.1007/s41870-022-01150-1}.
\bibitem[{Ngo et~al.(2024)Ngo, Gajula, Thorpe and Mckeever}]{NGO2024106558}
\bibinfo{author}{Ngo, V.M.}, \bibinfo{author}{Gajula, R.}, \bibinfo{author}{Thorpe, C.}, \bibinfo{author}{Mckeever, S.}, \bibinfo{year}{2024}.
\newblock \bibinfo{title}{Discovering child sexual abuse material creators' behaviors and preferences on the dark web}.
\newblock \bibinfo{journal}{Child Abuse \& Neglect} \bibinfo{volume}{147}, \bibinfo{pages}{106558}.
\newblock \DOIprefix\doi{10.1016/j.chiabu.2023.106558}.
\bibitem[{Ngo and Kechadi(2021)}]{Ngo:2021}
\bibinfo{author}{Ngo, V.M.}, \bibinfo{author}{Kechadi, M.T.}, \bibinfo{year}{2021}.
\newblock \bibinfo{title}{Electronic farming records – a framework for normalising agronomic knowledge discovery}.
\newblock \bibinfo{journal}{Computers and Electronics in Agriculture} \bibinfo{volume}{184}, \bibinfo{pages}{106074}.
\newblock \DOIprefix\doi{10.1016/j.compag.2021.106074}.
\bibitem[{Ngo et~al.(2019)Ngo, Le-Khac and Kechadi}]{ngo2019designing}
\bibinfo{author}{Ngo, V.M.}, \bibinfo{author}{Le-Khac, N.A.}, \bibinfo{author}{Kechadi, M.T.}, \bibinfo{year}{2019}.
\newblock \bibinfo{title}{Designing and implementing data warehouse for agricultural big data}, in: \bibinfo{booktitle}{International Conference on Big Data}, \bibinfo{organization}{Springer}. pp. \bibinfo{pages}{1--17}.
\bibitem[{Ngo et~al.(2020)Ngo, Le-Khac and Kechadi}]{ngo2020data}
\bibinfo{author}{Ngo, V.M.}, \bibinfo{author}{Le-Khac, N.A.}, \bibinfo{author}{Kechadi, M.T.}, \bibinfo{year}{2020}.
\newblock \bibinfo{title}{Data warehouse and decision support on integrated crop big data}.
\newblock \bibinfo{journal}{International Journal of Business Process Integration and Management} \bibinfo{volume}{10}, \bibinfo{pages}{17--28}.
\bibitem[{Ngo et~al.(2021)Ngo, Munnelly, Orlandi and Crooks}]{ngol2021semantic}
\bibinfo{author}{Ngo, V.M.}, \bibinfo{author}{Munnelly, G.}, \bibinfo{author}{Orlandi, F.}, \bibinfo{author}{Crooks, P.}, \bibinfo{year}{2021}.
\newblock \bibinfo{title}{A semantic search engine for historical handwritten document images}, in: \bibinfo{booktitle}{Linking Theory and Practice of Digital Libraries: 25th International Conference on Theory and Practice of Digital Libraries, TPDL 2021, Virtual Event, September 13--17, 2021, Proceedings}, \bibinfo{organization}{Springer Nature}. p.~\bibinfo{pages}{60}.
\bibitem[{Ngo et~al.(2025b)Ngo, Tran, Kearney and Roantree}]{ngo2025enhancing}
\bibinfo{author}{Ngo, V.M.}, \bibinfo{author}{Tran, V.Q.}, \bibinfo{author}{Kearney, P.}, \bibinfo{author}{Roantree, M.}, \bibinfo{year}{2025}b.
\newblock \bibinfo{title}{Enhancing bagging ensemble regression with data integration for time series-based diabetes prediction}, in: \bibinfo{booktitle}{Proceedings of the 17th International Conference on Computational Collective Intelligence (ICCCI'25)}, \bibinfo{publisher}{Springer}. pp. \bibinfo{pages}{1--15}.
\bibitem[{O'Brien et~al.(2016)O'Brien, Geoghegan, McNamara and Shalloo}]{Brien:16}
\bibinfo{author}{O'Brien, D.}, \bibinfo{author}{Geoghegan, A.}, \bibinfo{author}{McNamara, K.}, \bibinfo{author}{Shalloo, L.}, \bibinfo{year}{2016}.
\newblock \bibinfo{title}{How can grass-based dairy farmers reduce the carbon footprint of milk?}
\newblock \bibinfo{journal}{Animal Production Science} \bibinfo{volume}{56}, \bibinfo{pages}{495}.
\newblock \DOIprefix\doi{10.1071/an15490}.
\bibitem[{Pranga et~al.(2021)Pranga, Borra-Serrano, Aper, De~Swaef, Ghesquiere, Quataert, Roldán-Ruiz, Janssens, Ruysschaert, Lootens and et~al.}]{PBADGQ21}
\bibinfo{author}{Pranga, J.}, \bibinfo{author}{Borra-Serrano, I.}, \bibinfo{author}{Aper, J.}, \bibinfo{author}{De~Swaef, T.}, \bibinfo{author}{Ghesquiere, A.}, \bibinfo{author}{Quataert, P.}, \bibinfo{author}{Roldán-Ruiz, I.}, \bibinfo{author}{Janssens, I.A.}, \bibinfo{author}{Ruysschaert, G.}, \bibinfo{author}{Lootens, P.}, \bibinfo{author}{et~al.}, \bibinfo{year}{2021}.
\newblock \bibinfo{title}{Improving accuracy of herbage yield predictions in perennial ryegrass with uav-based structural and spectral data fusion and machine learning}.
\newblock \bibinfo{journal}{Remote Sensing} \bibinfo{volume}{13}, \bibinfo{pages}{3459}.
\newblock \DOIprefix\doi{10.3390/rs13173459}.
\bibitem[{Quan et~al.(2023)Quan, Lou, Lv, Sun, Xia, Li and Sun}]{Quan:2023}
\bibinfo{author}{Quan, L.}, \bibinfo{author}{Lou, Z.}, \bibinfo{author}{Lv, X.}, \bibinfo{author}{Sun, D.}, \bibinfo{author}{Xia, F.}, \bibinfo{author}{Li, H.}, \bibinfo{author}{Sun, W.}, \bibinfo{year}{2023}.
\newblock \bibinfo{title}{Multimodal remote sensing application for weed competition time series analysis in maize farmland ecosystems}.
\newblock \bibinfo{journal}{Journal of Environmental Management} \bibinfo{volume}{344}, \bibinfo{pages}{118376}.
\newblock \DOIprefix\doi{10.1016/j.jenvman.2023.118376}.
\bibitem[{Ruelle et~al.(2018)Ruelle, Hennessy and Delaby}]{RHD18}
\bibinfo{author}{Ruelle, E.}, \bibinfo{author}{Hennessy, D.}, \bibinfo{author}{Delaby, L.}, \bibinfo{year}{2018}.
\newblock \bibinfo{title}{Development of the moorepark st gilles grass growth model (most gg model): A predictive model for grass growth for pasture based systems}.
\newblock \bibinfo{journal}{European Journal of Agronomy} \bibinfo{volume}{99}, \bibinfo{pages}{80–91}.
\newblock \DOIprefix\doi{10.1016/j.eja.2018.06.010}.
\bibitem[{Sapkota et~al.(2020)Sapkota, Singh, Neely, Rajan and Bagavathiannan}]{Sapkota:2020}
\bibinfo{author}{Sapkota, B.}, \bibinfo{author}{Singh, V.}, \bibinfo{author}{Neely, C.}, \bibinfo{author}{Rajan, N.}, \bibinfo{author}{Bagavathiannan, M.}, \bibinfo{year}{2020}.
\newblock \bibinfo{title}{Detection of italian ryegrass in wheat and prediction of competitive interactions using remote-sensing and machine-learning techniques}.
\newblock \bibinfo{journal}{Remote Sensing} \bibinfo{volume}{12}.
\newblock \DOIprefix\doi{10.3390/rs12182977}.
\bibitem[{Tang et~al.(2023)Tang, Xu, Yang, Tang and Zhao}]{TANG2023103094}
\bibinfo{author}{Tang, C.}, \bibinfo{author}{Xu, L.}, \bibinfo{author}{Yang, B.}, \bibinfo{author}{Tang, Y.}, \bibinfo{author}{Zhao, D.}, \bibinfo{year}{2023}.
\newblock \bibinfo{title}{Gru-based interpretable multivariate time series anomaly detection in industrial control system}.
\newblock \bibinfo{journal}{Computers \& Security} \bibinfo{volume}{127}, \bibinfo{pages}{103094}.
\newblock \DOIprefix\doi{10.1016/j.cose.2023.103094}.
\bibitem[{Torres et~al.(2021)Torres, Hadjout, Sebaa, Martínez-Álvarez and Troncoso}]{THSMAT21}
\bibinfo{author}{Torres, J.F.}, \bibinfo{author}{Hadjout, D.}, \bibinfo{author}{Sebaa, A.}, \bibinfo{author}{Martínez-Álvarez, F.}, \bibinfo{author}{Troncoso, A.}, \bibinfo{year}{2021}.
\newblock \bibinfo{title}{Deep learning for time series forecasting: A survey}.
\newblock \bibinfo{journal}{Big Data} \bibinfo{volume}{9}, \bibinfo{pages}{3–21}.
\newblock \DOIprefix\doi{10.1089/big.2020.0159}.
\bibitem[{Viljanen et~al.(2018)Viljanen, Honkavaara, Näsi, Hakala, Niemeläinen and Kaivosoja}]{Viljanen:2018}
\bibinfo{author}{Viljanen, N.}, \bibinfo{author}{Honkavaara, E.}, \bibinfo{author}{Näsi, R.}, \bibinfo{author}{Hakala, T.}, \bibinfo{author}{Niemeläinen, O.}, \bibinfo{author}{Kaivosoja, J.}, \bibinfo{year}{2018}.
\newblock \bibinfo{title}{A novel machine learning method for estimating biomass of grass swards using a photogrammetric canopy height model, images and vegetation indices captured by a drone}.
\newblock \bibinfo{journal}{Agriculture} \bibinfo{volume}{8}.
\newblock \DOIprefix\doi{10.3390/agriculture8050070}.
\bibitem[{Wu and Kumar~Sangaiah(2021)}]{WS21}
\bibinfo{author}{Wu, H.}, \bibinfo{author}{Kumar~Sangaiah, A.}, \bibinfo{year}{2021}.
\newblock \bibinfo{title}{Oral english speech recognition based on enhanced temporal convolutional network}.
\newblock \bibinfo{journal}{Intelligent Automation \& Soft Computing} \bibinfo{volume}{28}, \bibinfo{pages}{121–132}.
\newblock \DOIprefix\doi{10.32604/iasc.2021.016457}.
\bibitem[{Yoo and Oh(2020)}]{Yoo:2020}
\bibinfo{author}{Yoo, T.W.}, \bibinfo{author}{Oh, I.S.}, \bibinfo{year}{2020}.
\newblock \bibinfo{title}{Time series forecasting of agricultural products’ sales volumes based on seasonal long short-term memory}.
\newblock \bibinfo{journal}{Applied Sciences} \bibinfo{volume}{10}.
\newblock \DOIprefix\doi{10.3390/app10228169}.
\bibitem[{Yuan et~al.(2023)Yuan, Lin, Zhou, Jiang and Liu}]{Yuan:2023}
\bibinfo{author}{Yuan, Y.}, \bibinfo{author}{Lin, L.}, \bibinfo{author}{Zhou, Z.G.}, \bibinfo{author}{Jiang, H.}, \bibinfo{author}{Liu, Q.}, \bibinfo{year}{2023}.
\newblock \bibinfo{title}{Bridging optical and sar satellite image time series via contrastive feature extraction for crop classification}.
\newblock \bibinfo{journal}{ISPRS Journal of Photogrammetry and Remote Sensing} \bibinfo{volume}{195}, \bibinfo{pages}{222--232}.
\newblock \DOIprefix\doi{10.1016/j.isprsjprs.2022.11.020}.
\bibitem[{Zhang et~al.(2022)Zhang, Song, Chen, Wang, Wang and Li}]{ZSCWWL22}
\bibinfo{author}{Zhang, R.}, \bibinfo{author}{Song, H.}, \bibinfo{author}{Chen, Q.}, \bibinfo{author}{Wang, Y.}, \bibinfo{author}{Wang, S.}, \bibinfo{author}{Li, Y.}, \bibinfo{year}{2022}.
\newblock \bibinfo{title}{Comparison of arima and lstm for prediction of hemorrhagic fever at different time scales in china}.
\newblock \bibinfo{journal}{PLOS ONE} \bibinfo{volume}{17}.
\newblock \DOIprefix\doi{10.1371/journal.pone.0262009}.
\bibitem[{Zhou and Soldat(2021)}]{ZS21}
\bibinfo{author}{Zhou, Q.}, \bibinfo{author}{Soldat, D.J.}, \bibinfo{year}{2021}.
\newblock \bibinfo{title}{Creeping bentgrass yield prediction with machine learning models}.
\newblock \bibinfo{journal}{Frontiers in Plant Science} \bibinfo{volume}{12}.
\newblock \DOIprefix\doi{10.3389/fpls.2021.749854}.

\end{thebibliography}

\end{document}